# DID WILLIAM SHAKESPEARE AND THOMAS KYD WRITE *EDWARD III*?

David Kernot[1, 2], Terry Bossomaier[3] and Roger Bradbury[1]

[1] National Security College. The Australian National University, Canberra, ACT, Australia.
[2] Defence Science & Technology Group, Edinburgh, SA, Australia.
[3] The Centre for Research in Complex Systems. Charles Sturt University, Bathurst, NSW, Australia.

*ABSTRACT*

*William Shakespeare is believed to be a significant author in the anonymous play, The Reign of King Edward III, published in 1596. However, recently, Thomas Kyd, has been suggested as the primary author. Using a neurolinguistics approach to authorship identification we use a four-feature technique, RPAS, to convert the 19 scenes in Edward III into a multi-dimensional vector. Three complementary analytical techniques are applied to cluster the data and reduce single technique bias before an alternate method, seriation, is used to measure the distances between clusters and test the strength of the connections. We find the multivariate techniques robust and able to allocate up to 14 scenes to Thomas Kyd, and further question if scenes long believed to be Shakespeare's are not his.*

## 1. INTRODUCTION

*The Reign of King Edward III* (*Edward III*) was first published in 1596 and is of uncertain authorship (Slater, 1988). However, it is a new addition to the Shakespeare canon, and even while there is a suggestion that Shakespeare is not the sole author, he is considered to be a significant one (Shakespeare, & Melchiori, 1998).The idea that the work might be Shakespeare's was first suggested by Edward Capell in 1760 (Champion, 1988). Many others have offered their own candidate from the list of popular playwrights of the time since then, including that the 19 scenes within itare all Shakespeare's, but more recently, Brian Vickers, using plagiarism detection software, has suggested that Thomas Kyd is the major author with Shakespeare having a lesser role (Vickers, 2014). Using techniques that have successfully identified the works of William Shakespeare, Christopher Marlowe, and several other Elizabethan playwrights and poets (Kernot, Bossomaier, & Bradbury, 2017a; 2017b), we test the claim that the anonymous play, *Edward III,* was co-written by William Shakespeare and Thomas Kyd.

### 1.1 AN APPROACH

While many text analysis techniques rely on basic statistical correlations, word counts, collocated word groups, or keyword density (Lamb, Paul, & Dredze, 2013; Leech &On wuegbuzie, 2007; Matsuo & Ishizuka, 2004), Vicker's approach using trigrams, instances where three consecutive words in a sentence closely match known authored works, is a more recent technique.

In this paper, we employ a new methodology that adopts a multi-faceted approach to text analysis and reveal details about a person's personality; their sense of self, from subtle characteristics hidden in their writing style (Argamon *et al.,* 2009; Iqbal *et al.*, 2013; Northoff *et al.*, 2006). We use four neurolinguistic indicators, known as **RPAS** (Kernot, Bossomaier, & Bradbury, 2016). They draw on biomarkers for creativity and known psychosis (Rosenstein *et al.*, 2015; Zabelina *et al.*, 2015) to identify characteristics within the *Edward III* scenes. **RPAS** is used to to create a stylistic signature from a person's writing: **Richness (R)** (Menhinick, 1964; Tweedie, &Baayen, 1998), the number of unique words used by an author and linked to education and age (Hartshorne & Germine, 2015); **Personal Pronouns (P)** (Argamon *et al.*, 2003; Kernot, 2016; Pennebaker, 2011; Pennebaker, Mehl, & Niederhoffer, 2003), the pronouns used, closely aligned to self and a person's socially constructed or internal gender style; **Referential Activity Power (A)** (Bucci, 2002 ; Bucci & Maskit, 2004), based on function words, or word particles derived from clinical depression studies; and **Sensory (S)** (Kernot, 2013; Lynott, & Connell, 2009; Miller, 1995; van Dantzig *et al.*, 2011; Fernandino *et al.*, 2015), five sensory measures (V-Visual A-Auditory H – Haptic O – Olfactory G - Gustatory) corresponding to the use of the senses.

We use **RPAS** to create stylistic signatures of the known works of Shakespeare, Kyd, and Marlowe and compare them to the 19 scenes within the *Edward III* play to suggest the authorship. We also label the four commonly understood scenes attributed to William Shakespeare (scenes 1.2, 2.1, 2.2. and 4.4 which we refer to as chunks 2, 3, 4 and 13).We then analyse and cluster the results to identify the likely authorship of the 19 unknown scenes within *Edward III*.

## 2. METHODOLOGY

We draw on the June 1999 Project Gutenberg Etext of *The Reign of King Edward the Third*, attributed in part to William Shakespeare, and the February 2011 Project Gutenberg EBook of *The Spanish Tragedy*, by Thomas Kyd. While scholars have argued that Shakespeare's writing can be seen in the additional passages of Thomas Kyd's fourth quarto of *The Spanish Tragedy (*Bruster, 2013), we have used an earlier version, the second quarto printed in 1592 to avoid any influence of Shakespeare in the results. Consideration was given to using additional works from Kyd. *Cornelia* was discarded because it is a translation of a known earlier work of another author. The anonymous play *Soliman and Perseda* is now being attributed to Kyd because it is presented as a summarized plot in *The Spanish Tragedy*, however features within the play have also been attributed to Shakespeare, Marlowe and Kyd (Merriam, 1995) and it too was discarded in favour of *The Spanish Tragedy.*

We remove all play stage direction and process both files with the Stanford Parts Of Speech Tagger (Toutanova & Manning, 2000). Punctuation is removed and then we aggregate the works by word frequency. *The Reign of King Edward the Third* is further broken down into segments marked by the 19 scenes. The average scene size is 1035 words (88 – 3720) and the RPAS technique relies on normalizing the results based on sample size. Words as small as a sonnet (91-132 words) have been used successfully with this approach and the upward file size limit has been calculated at 10,000 words (see Kernot, Bossomaier, & Bradbury, 2016).

We use William Shakespeare's *Venus and Adonis*, and Christopher Marlowe's *Hero and Leander,* which are drawn from a pre-processed corpus (Kernot, Bossomaier, & Bradbury, 2016) and originally sourced from the complete works of Shakespeare (Farrow, 1993) and Marlowe (Farey, 2014). We norrmalise the RPAS Personal Pronouns (P) across the 21,300 words from Kyd's *The Spanish Tragedy*, the 19,600 words in the anonymous *Edward III* play, and divide them by the 897,000 word corpus of Shakespeare and Marlowe so that all of the Personal Pronoun results are consistent across all tests.

We create a nine-dimensional array from the data using RPAS and apply three complementary techniques to reduce any single bias and determine the possible authorship of the 19 scenes. As a final measure, we then use seriation to visualise the nine-dimensional array as a one-dimensional continuum and get some distance metrics between the clusters, before adding noise to test the strength of the co-located cluster edges.

## 2.1 THREE COMPLEMENTARY TECHNIQUES

Apart from examining the individual RPAS values from the 19 *Edward III* scenes (which includes the Sensory sub-elements VAHOG Visual, Auditory, Haptic, Olfactory, and Gustatory measures), we plot Personal Pronouns (P) against Richness (R) (PtoR), Referential Activity Power (A) to Richness (R) (AtoR), and Sensory Adjectives (S) to Richness (R) (StoR) and examine the clusters that form.

We then use the Vector Space Method (VSM) technique (Koppel, & Winter, 2014; Voorhees, 1998). We conduct pair-wise comparisons of each of the 19 *Edward III* scenes against Thomas Kyd's work, *The Spanish Tragedy* and William Shakespeare's poem *Venus and Adonis* (each a 38 pair-wise comparison) using both cosine and minmax similarity detection (Koppel & Seidman, 2013). and plot them to examine the clusters.

Extending the cosine and minmax approach of VSM, we then use the imposters method (Seidman, 2013), where we compare work that is *not* the work of either of the two authors and in this case is used to cluster commonly authored scenes. This method gives surprisingly strong results for the verification problem, even when the documents in question contain no more than 500 words (Koppel & Winter, 2014). We select Christopher Marlowe's poem, *Hero and Leander,* completed by George Chapman after Marlowe's death which is very stylistically different to both Kyd and Shakespeare's work. This technique has been used to separate the different authors in the Elizabethan publication, The Passionate Pilgrim (Kernot, Bossamier & Bradbury, 2017a).

## 2.2 SERIATION

According to Liiv (2010:71) "Seriation is an exploratory combinatorial data analysis technique to reorder objects into a sequence along a one-dimensional continuum so that it best reveals regularity and patterning among the whole series." Seriation is the process of placing a linear ordering on a set of N multi-dimensional quantities. The total number of possible orderings is N! (factorial). This grows extremely quickly with N. $5! = 120$, $10! = 3.6$million and $20! = 2.4 \times 10^{18}$, or 2.4 billion billion(or quintillion). Thus even for quite small N, we can't calculate the shortest path by calculating all possible paths. We need a heuristic or approximation. Inevitably any given approximation will work better with some data than others. Thus, for a robust estimation of the shortest path it might be necessary to try a range of different estimators and look for consistency among them.

We use the free software environment for statistical computing and graphics, R, and its seriation package (Buchta, Hornik, & Hahsler, 2008), and provide the seriation package with the 9x19 matrix consisting of the nine RPAS values for each of the 19 scenes of *Edward III*. Using the Euclidean distance option, seriation attempts to minimise the Hamiltonian path length (the Hamiltonian path on a graph is a path which visits all the nodes just once).We evaluated the results of the six Hamiltonian path-length calculations produced by the seriation package (TSP: *Travelling Salesperson*, Chen: *Rank two ellipse Seriation,* ARSA: *Anti-Robinson Simulated Annealing*, HC: *Hierarchical Clustering*, GW: *Hierarchical Clustering (GruvaeusWainer heuristic)*, and OLO: *Hierarchical Clustering (Optimal Leaf Ordering))*. While seriation gives a one-dimensional continuum, Dendrogram branch and leaf visualization is also provided, and clusters can be separated by their Hamiltonian path distances (Earle, & Hurley, 2015). We select the technique that provides the shortest Hamiltonian path and introduce noise into the matrix to examine the strength of the connected groups by using the jitter function in R. The function adds random noise to the vector by drawing samples from the uniform distribution of the original data (Stahel & Maechler, 2011).

## 3. RESULTS

We analyze the 19 *Edward III* scenes using **RPAS** and visual indications of the nine variables highlight variation in Richness (R), Personal Pronouns (P) and the five Sensory (S) elements (VAHOG). Discriminating by sensory-based VAHOG measures having zero scores, chunks 8, 10, 11, 12, 14, 15, 16, 17stand out, and we include chunks 6 and 18 to the list when we consider Richness and Personal Pronouns scores (R >50 or P > 0).

### INSERT FIGURE 1

We conduct visual clustering and plot Personal Pronouns (P) against Richness (R) (P to R), Referential Activity Power (A) against Richness (R) (A to R), and Sensory Adjectives (S) against Richness (R) (S to R) and examine the results (see Figure1 (P to R) and Figure2 (S to R). A to R is omitted because it mimics P to R).P to R discriminates chunks 6, 8, 10, 11, 15, 16, 17, and 18 by Richness and Personal Pronouns. Of these, chunks 6, 8, 15, 16, 17 have a richer and less feminine internal gendered style (R >50 and P >8) while chunks 10, 11, and 18 appear ambiguous. A to R reinforces the Richness aspects but does not contribute further. S to R supports the P to R results and discriminates chunks 3, 6, 8, 10, 12, 15, 16, 17, and 18 by Richness and a wide sensory range. Of these, chunks 6, 8, 15, 16, and 17 have a richer and much wider sensory range (where R>50), while chunks 10, 12, and 18 appear ambiguous.

### INSERT FIGURE 2

The cluster assignment is discretionary, however, through amalgamation these techniques identify two groups(Group 1: 1, 2, 3, 4, 5, 7, 9, 13, 14, 19 and Group 2: 6, 8, 10, 11, 12, 15, 16, 17, 18)with some variation in chunks 10, 11, 12, and 18.

### 3.1 VECTOR SPACE METHOD (VSM)

We create a stylistic signature of Thomas Kyd's play, *The Spanish Tragedy* using the **RPAS** method, and compare it to the 19 *Edward III* scenes using both cosine and minmax similarity detection (38 pair-wise comparisons with nine dimensions). We plot these as an XY Cartesian product in Figure 3 and examine the clusters. We expect Kyd's authored chunks to appear in the upper-right-hand corner, and ones furthest away (bottom-left-hand corner) to be Shakespeare's works. The similarity plot (Figure 3) highlights twoclusters, and we assign Kyd's authorship toone ( chunks 1, 2, 5, 7, 9, 10, 11, 12, 13, 14, 18, 19 ) and Shakespeare to the other ( chunks 6, 8, 15, 16, 17 ).Chunks 3 and 4 sit outside but also indicate Kyd.

### INSERT FIGURE 3

Next, we compare Shakespeare's *Venus and Adonis* to Thomas Kyd's *The Spanish Tragedy* using both cosine and minmax similarity detection. The similarity between Shakespeare and Kyd's work is small (cosine 21.96%, minmax 18.29% or~79.9% dissimilar) so Shakespeare's authored chunks will sit closer to the upper-right-hand corner. The similarity plot (Figure 4) highlights a Shakespeare cluster (chunks 6, 8, 11, 15, 16, 17), and a Kyd cluster (chunks 1, 2, 4, 5, 7, 9, 12, 13, 14, 18, 19). Chunks 3 and10 sit outside and are more like Kyd on the cosine measure and ambiguous on the minmax measure. They also correlate to Figure 3,.

INSERT FIGURE 4

## 3.2 IMPOSTERS METHOD

We compare Christopher Marlowe's *Hero and Leander* to Thomas Kyd's *The Spanish Tragedy* using both cosine and minmax. Marlowe's work is considered to an imposter because he is neither Kyd or Shakespeare. The similarity between Marlowe and Kyd's work is small(cosine 22.096%, minmax 17.48% or ~80.2% dissimilar) so Kyd's authored chunks will sit furthest from the upper-right-hand corner. The similarity plot (Figure 5) highlights a Shakespeare cluster (chunks 6, 8, 11, 15, 16, 17), and a Kyd cluster (chunks 1, 2, 4, 5, 7, 9, 12, 13, 14, 18, 19). Chunks 3 and 10 sit outside and are more like Kyd on the cosine measure, but ambiguous on the minmax measure.

INSERT FIGURE 5

## 3.3 VSM USING THE SPANISH TRAGEDY CHUNKS

There are four commonly accepted scenes attributed to William Shakespeare by scholars (clusters 2, 3, 4, and 13 are red circles in the figures) and in all cases we find they sit inside or close to Kyd's clusters. We conduct further VSM analysis to counter the commonly held view that these are Shakespeare's.

We chunk Thomas Kyd's *The Spanish Tragedy* into 25 scenes and conduct pair-wise comparisons of each of them against the four *Edward III* scenes attributed to Shakespeare (100 pair-wise comparisons) using both cosine and minmax similarity detection. We find 13 of the 25 scene chunks clearly identify with Kyd (~52%), which we believe is a relatively high number given the smaller size comparisons.

## 3.4 SERIATION

Throughout these different visualization techniques we see variability with chunks 6, 10, 11, and 18, and the four commonly accepted scenes attributed to William Shakespeare (chunks 2, 3, 4, and 13) identify as Kyd, but many of the techniques have been dependent on an arbitrary visual clustering size. Therefore to add further reliability to the results, we cluster the data using seriation.

The R seriation package is fed a 9x19 matrix of the data, and using Euclidean distance we seriate the data to minimize the Hamiltonian path length. Results of the six seriation techniques available highlight that Hierarchical Clustering with Optimal Leaf Ordering (OLO) outperforms the Travelling Salesperson technique (path lengths 140.96 vs. 157.52).The order of the 19 chunks is 8 17 15 16 6 10 9 11 18 19 13 14 12 5 1 2 7 4 3 (see Table 1 for more detail). When we compare the distances between each chunk the ordering sequence is important, but the distance information does not convey much.

## INSERT TABLE1

To see how stable the results are, we insert noise into the initial 9x19 RPAS-scene matrix and re calculate the Euclidean distances with various amounts of noise (between 1 – 2000).  An examination of the scene chunk order after seriation (see Table  2) highlights the susceptibility of chunks 10 and 11 to moderate amounts of noise, and there is some movement of the order of the scene chunks in the middle section with a significant amount of introduced noise.  However, we find the Shakespeare chunks (6, 8, 15, 16, 17) do not move, nor do the three chunks commonly attributed to Shakespeare (2, 3, and 4), which remain in a large group with Kyd's work (chunks 1, 2, 3,4,5, 7).

**INSERT TABLE2**

## 4. DISCUSSION

When we examine the 19 *Edward III* scenes using **RPAS,** the Visual, Auditory, Haptic, Olfactory, and Gustatory (VAHOG) Sensory (S) results split the data into two distinct groups (a 57/42% split). The data can also be separated by Richness (R) and Personal Pronouns (P) with similar results.The amalgamation of the PtoR, AtoR, and StoR analysis clusters chunks 6, 8, 15, 16, and 17. These chunks have a richer and much wider sensory range with a lesser feminine style (where $R > 50$ and $P > 8$), and it highlights the significance of using Richness, Personal Pronouns, and the Sensory VAHOG variables in RPAS. Chunks 10, 11, 12, and 18 stand out but appear to be ambiguous.

By using VSM we can compare *Edward III* to a known work of Thomas Kyd, and we assign chunks 6, 8, 15, 16, and 17 to Shakespeare and the rest to Kyd. Using Shakespeare's *Venus and Adonis*, we again see some further variability in chunk 10 and 11, but overall the chunks are consistent with the previous techniques. These results are reflected in the Imposters Method with VSM using Marlowe's *Hero and Leander*. The only change from Shakespeare's is the order of chunks 6 and 8, and again this reinforces the overall results adding another layer of consistency. Using the imposter method in a study of 42 commonly attributed works of Shakespeare that also included both plays of Thomas Kyd and the Edward III play, Koppel and Winter (2014) findings suggest that Edward III is more similar to Thomas Kyd's plays than 39 of Shakespeare's.

We also find all four commonly attributed Shakespeare scenes (chunks 2, 3, 4, and 13) consistently fall inside the Kyd clusters or close to them. Using 'new-optics' stylometric measures on Edward III play, Elliot and Valenza (2010) findings suggest that when taken as a group, Shakespeare's authorship of the four scenes commonly attributed to him are unlikely, a view they say is supported in Marina Tarlinskaja (2006) unpublished article.

When we conducted further VSM analysis by using VSM and chunking Thomas Kyd's *The Spanish Tragedy* into 25 scenes shows close similarities to 52% of Kyd's work. The results we believe, given the small size of the chunks, would appear to be a relatively high number of similar works and supports the earlier results that their authorship is likely Kyd's and not Shakespeare's.

We find the arbitrary nature of the clustering size does influence the reporting to a small degree, and while we believe the cluster sizes reasonable, there has been some minor variability with chunks 6 and 18, but more so with chunks 10 and 11. However, using Seriation, it is clear that chunk 6 is part of the 'Shakespeare 5' (sits alongside chunks 8, 15, 16, and 17). Chunks 10, 9, 11, and 18 are the closest chunks to the Shakespeare cluster but are separate from him. By adding a moderate amount of noise to the seriation matrix, we find some variability with chunks 10 and 11. It is possible that they are collaborative scenes containing both the work of Kyd and Shakespeare. However, of the commonly accepted Shakespeare scenes, three of them clustered together (chunks 2, 3, and 4) at the opposite end to the Shakespeare work and no amount of introduced noise moved or separated them from Kyd's work. Only chunk 13 sits away, and while it is six scenes from the 'Shakespeare 5', it is closer to Kyd.

In comparing these results to Vickers' (2014), we find we agree with nine of the scenes that he has suggested are Kyd's, and this analysis suggests that scenes4.1 (chunk 10) and 4.2 (chunk 11) appear to be Kyd Shakespeare collaborations. We disagree with his analysis of scenes 3.2, 3.4, 4.6, and 4.8 (chunks 6, 8, 15, and 17 from the 'Shakespeare 5' cluster). We also suggest that the four scenes commonly attributed to Shakespeare, scenes 1.2, 2.1, 2.2, and 4.4 (chunks 2, 3, 4, and 13) are written by Kyd, although scene 2.1 and to a lesser extent scene 2.2 are more 'Kyd-like' and away from the main body of the other Kyd scenes (see Elliot & Valenza, 2010 for similarities to these findings). As we show in Table 3, we agree with Vickers' conclusion that the majority of the work *Edward III* was written by Kyd.

Using **RPAS**, we identify subtle characteristics within Shakespeare that identify him separately to Kyd. Shakespeare uses more unique words and less repetition than Kyd, less feminine personal pronouns and more masculine ones, and overall he used a wider range of visual descriptions, but draws less on sensory characteristics and emotional experiences, and is vaguer and more general than Kyd.

### 4.1. A LIMITATION OF THE OVERALL APPROACH

A limitation of this overall approach is that the results are dependent on the chunking of the data into 19 scenes. Elliot and Valenza (2010) split scene 2.1 into two parts. If these scenes are not the true delineation between the efforts of two authors, then this would skew the results, but at the end of the day, it is difficult to tell what, if any, a split in the scenes may have been. Here we assume that each scene was written by a single author. However, if this were not true then that scene would appear stylistically different from both William Shakespeare and Thomas Kyd's other works as a third author, and this did not occur, although as we have stated, scene 2.1 and to a lesser extent scene 2.2 were more 'Kyd-like' than the other scenes, and there were two works that were similar to both authors (chunks 10 and 11, or scenes 4.1 and 4.2) and they could well be collaborations. In dealing with over 400 years old text, we suggest the exact details and events that led to this fascinating union of work by Thomas Kyd and William Shakespeare may well be lost within the strands of time.

INSERT TABLE 3

### 5. CONCLUSIONS

In this analysis, the four scenes commonly attributed to Shakespeare identify as Thomas Kyd, but this is not unexpected (see Elliot & Valenza, 2010). However, it seems clear from analysis that Thomas Kyd wrote the majority of the play and William Shakespeare played a lesser role. On the basis of these findings, the collaborative play, *The Reign of King Edward III,* could well have been written by William Shakespeare and Thomas Kyd.

In examining this multivariate technique, we find the analysis provided a consistent result and therefore the techniques were resilient. The results of seriation were found to be robust to perturbations in the **RPAS** features and strongly validate the approach to author identification. Significant differentiation was found using RPAS and the neuro linguistics approach of Richness (R), internal or socially constructed, gendered Personal Pronouns (P), Referential Activity power (A), and Sensory Adjectives(S).

### ACKNOWLEDGMENTS

We thank T. Pattison and A. Ceglar for reading an early version of this manuscript and providing critical comments. This research is supported by the Defence Science Technology Group, the Australian Government's lead agency dedicated to providing science and technology support for the country's defence and security needs.

**Funding:** This research did not receive any specific grant from funding agencies in the public, commercial, or not-for-profit sectors.

**REFERENCES**

[1] Argamon, S., Koppel, M., Fine, J., Shimoni, A.R. (2003). Gender, genre, and Writing Style in Formal Written Texts. Text, Volume 23, Number 58, August 2003.
[2] Argamon, S., Koppel, M., Pennebaker, J. W., &Schler, J. (2009).Automatically profiling the author of an anonymous text.Communications of the ACM, 52(2), 119-123.
[3] Bruster, D. (2013). Shakespearean spellings and handwriting in the additional passages printed in the 1602 Spanish Tragedy. Notes and Queries, gjt124.
[4] Bucci, W. (2002).The referential process, consciousness, and the sense of self.Psychoanalytic Inquiry, 22(5), 766-793.
[5] Bucci, W., &Maskit, B. (2004).Building a weighted dictionary for referential activity.In Spring Symposium of the American Association for Artificial Intelligence in Palo Alto, CA, March.
[6] Buchta, C., Hornik, K., &Hahsler, M. (2008).Getting things in order: an introduction to the R package seriation.Journal of Statistical Software, 25(3), 1-34.
[7] Champion, L. S. (1988). 'Answere to this perillous time': Ideological ambivalence in the raigne of king Edward III and the english Chronicle plays.
[8] Earle, D., & Hurley, C. B. (2015).Advances in dendrogram seriation for application to visualization.Journal of Computational and Graphical Statistics, 24(1), 1-25.
[9] Elliott, W. E., &Valenza, R. J. (2010). Two tough nuts to crack: did Shakespeare write the 'Shakespeare' portions of Sir Thomas More and Edward III? Part I. Literary and linguistic computing, 25(1), 67-83.
[10] Farey, P. (2014).Peter Farey's Marlowe Pagehttp://www2.prestel.co.uk/
[11] Farrow, J. M. (1993) The Collected Works of Shakespeare. http://sydney.edu.au/engineering/it/~matty/Shakespeare/
[12] Fernandino, L., Binder, J. R., Desai, R. H., Pendl, S. L., Humphries, C. J., Gross, W. L., ... & Seidenberg, M. S. (2015). Concept Representation Reflects Multimodal Abstraction: A Framework for Embodied Semantics. Cerebral Cortex, bhv020.
[13] Hartshorne, J. K., &Germine, L. T. (2015). When Does Cognitive Functioning Peak? The Asynchronous Rise and Fall of Different Cognitive Abilities Across the Life Span. Psychological science, 09567976145673039
[14] Iqbal, F., Binsalleeh, H., Fung, B., &Debbabi, M. (2013).A unified data mining solution for authorship analysis in anonymous textual communications.Information Sciences, 231, 98-112.
[15] Kernot, D. (2013). The Identification of Authors Using Cross-Document Co-Referencing http://www.unsworks.unsw.edu.au/primo_library/libweb/action/dlDisplay.do?vid=UNSWORKS&docId=unsworks_12072.
[16] Kernot, D. (2016) Can Three Pronouns Discriminate Identity in Writing in Data. In Sarker, R., Abbas, H., Dunstall, S., Kilby, P., Davis, R. Young, L. (eds) Data and Decision Sciences in Action: Proceedings of the Australian Society for Operations Research Conference 2016, Springer.
[17] Kernot, D., Bossomaier, T., & Bradbury, R. (2017a).Stylometric Techniques for Multiple Author Clustering: Shakespeare's Authorship in The Passionate Pilgrim. International Journal of Advanced Computer Science and Applications. (Accepted 14 March 2017 for Vol. 8 No. 3 March 2017).
[18] Kernot, D., Bossomaier, T., & Bradbury, R. (2017b). Novel Text Analysis for Investigating Personality: Identifying the Dark Lady in Shakespeare's Sonnets (Journal of Quantitative Linguistics – accepted 18 Jan 2017).
[19] Koppel, M., & Seidman, S. (2013).Automatically Identifying Pseudepigraphic Texts.In EMNLP (pp. 1449-1454).
[20] Koppel, M., &Winter, Y. (2014). Determining if two documents are written by the same author.Journal of the Association for Information Science and Technology, 65(1), 178-187.
[21] Lamb, A., Paul, M. J., &Dredze, M. (2013, June).Separating Fact from Fear: Tracking Flu Infections on Twitter.In HLT-NAACL (pp. 789-795).
[22] Leech, N. L., &Onwuegbuzie, A. J. (2007). An array of qualitative data analysis tools: A call for data analysis triangulation. School psychology quarterly, 22(4), 557.
[23] Liiv, I. (2010). Seriation and matrix reordering methods: An historical overview. Statistical analysis and data mining, 3(2), 70-91.
[24] Lynott, D., & Connell, L. (2009).Modality exclusivity norms for 423 object properties. Behavior Research Methods, 41(2), 558-564.

[25] Matsuo, Y., & Ishizuka, M. (2004).Keyword extraction from a single document using word co-occurrence statistical information.International Journal on Artificial Intelligence Tools, 13(01), 157-169.
[26] Menhinick, E. F. (1964). A comparison of some species-individuals diversity indices applied to samples of field insects. Ecology, 859-861.
[27] Merriam, T. (1995).Possible light on a Kyd canon.Notes and Queries, 42(3), 340-341.
[28] Miller, G. A. (1995).The science of words. New York: Scientific American Library.
[29] Northoff, G., Heinzel, A., de Greck, M., Bermpohl, F., Dobrowolny, H., &Panksepp, J. (2006).Self-referential processing in our brain—a meta-analysis of imaging studies on the self. Neuroimage, 31(1), 440-457.
[30] ]Pennebaker, J. W. (2011). The secret life of pronouns.New Scientist, 211(2828), 42-45.
[31] Pennebaker, J. W., Mehl, M. R., &Niederhoffer, K. G. (2003).Psychological aspects of natural language use: Our words, our selves. Annual review of psychology, 54(1), 547-577.
[32] Rosenstein, M., Foltz, P. W., DeLisi, L. E., &Elvevåg, B. (2015).Language as a biomarker in those at high-risk for psychosis.Schizophrenia research.
[33] Seidman, S. (2013). Authorship verification using the impostors method. In CLEF 2013 Evaluation Labs and Workshop-Online Working Notes.
[34] Shakespeare, W., &Melchiori, G. (1998).King Edward III. Cambridge University Press.
[35] Slater, E. (1988). The Problem of The Reign of King Edward III: A Statistical Approach. Cambridge University Press.
[36] Stahel, W., Maechler, M. (2011). 'Jitter' (Add Noise) to Numbers. R Documentation (1995 – 2011) available at: http://stat.ethz.ch/R-manual/R-devel/library/base/html/jitter.html. Accessed: 2 August 2016.
[37] Tarlinskaja, M. (2006) Shakespeare Among Others in Edward III and Sir Thomas More: From Meter to Authorship. Seattle, Washington.
[38] Toutanova, K., & Manning, C. D. (2000, October). Enriching the knowledge sources used in a maximum entropy part-of-speech tagger. In Proceedings of the 2000 Joint SIGDAT conference on Empirical methods in natural language processing and very large corpora: held in conjunction with the 38th Annual Meeting of the Association for Computational Linguistics-Volume 13 (pp. 63-70). Association for Computational Linguistics.
[39] Tweedie, F. J., &Baayen, R. H. (1998).How variable may a constant be? Measures of lexical Richness in perspective.Computers and the Humanities, 32(5), 323-352.
[40] VanDantzig, S., Cowell, R. A., Zeelenberg, R., &Pecher, D. (2011). A sharp image or a sharp knife: Norms for the modality-exclusivity of 774 concept-property items. Behavior research methods, 43(1), 145-154
[41] Vickers, B. (2014). "The Two Authors of Edward III",Shakespeare Survey.Ed. Peter Holland.1st ed. Vol. 67. Cambridge: Cambridge University Press, 2014. pp. 102-118. Shakespeare Survey Online.Web. 05 March 2015. http://dx.doi.org.virtual.anu.edu.au/10.1017/SSO9781107775572.008
[42] Voorhees, E. M. (1998). Using WordNet for text retrieval.Fellbaum (Fellbaum, 1998), 285-303.
[43] Zabelina, D. L., O'Leary, D., Pornpattananangkul, N., Nusslock, R., &Beeman, M. (2015).Creativity and sensory gating indexed by the P50: Selective versus leaky sensory gating in divergent thinkers and creative achievers. Neuropsychologia, 69, 77-84.

Figure 1: In this *Edward III* gendered Personal pronouns (P) versus Richness(R) diagram we see chunks 6, 8, 10, 11, 15, 16, 17, and 18 with greater than 50% richness or greater than 0 gendered personal pronouns. Of these, we see the 'Shakespeare 5' in the smaller shaded ellipse having higher female gender scores and Richness, while 18, 10, and 11 might be ambiguous. Of note, the four Shakespearian clusters marked with a red circle are those commonly attributed to Shakespeare. Further, the ellipses are our visual clustering assignment.

International Journal on Natural Language Computing (IJNLC) Vol. 6, No.6, December 2017

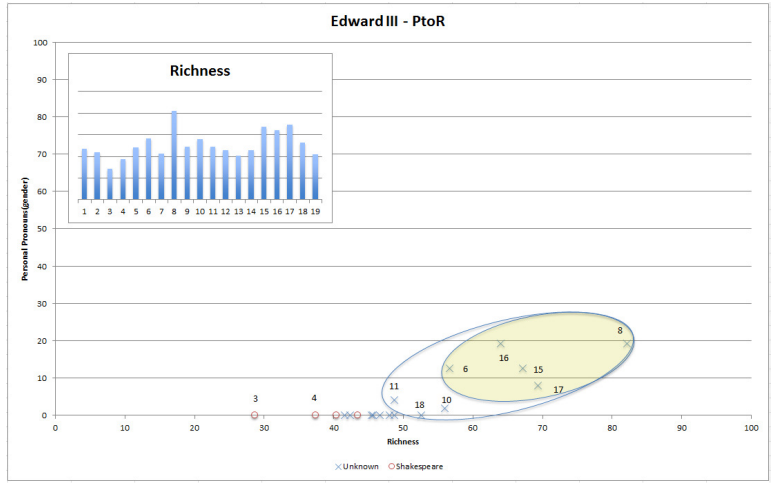

Figure 2: In this Edward III Sensory Adjectives(S) versus Richness(R) diagram with the VAHOG sensory allocations, we see chunks 3, 6, 8, 10, 12,15, 16, 17, and 18 stand out because Richness is greater than 50% or they have a wide sensory range when compared to the shaded circle Kyd chunks. Of these, we see the 'Shakespeare 5' (6, 8, 15, 16, and 17) having a much wider sensory range with higher Richness, while 18, and 10 might be ambiguous. Of note, the four Shakespearian clusters marked with a red circle are those commonly attributed to Shakespeare. Further, the shaded circle is our visual clustering assignment.

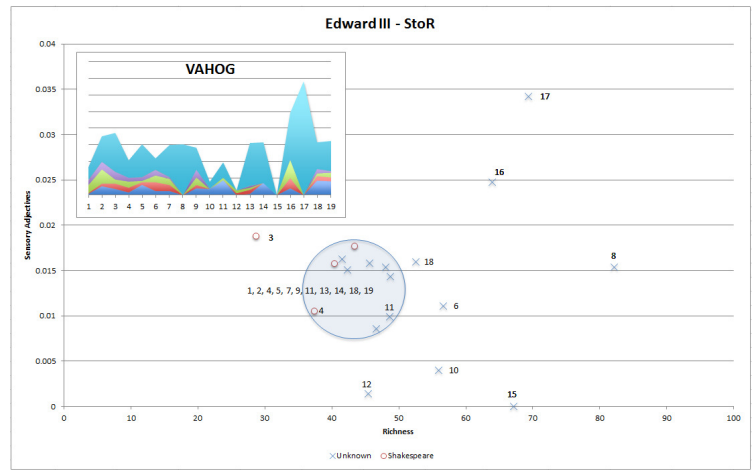

Figure 3: Using the 9-dimensional RPAS vector we compare Thomas Kyd's Spanish Tragedy to Edward III scene chunks using min max and cosine similarity detection. We see the extreme values of chunks 3 and 4(commonly attributed to Shakespeare) and these are clearly Kyd on this metric. The 'Shakespeare 5' (chunks 6, 8, 15, 16, 17)appears in the lower cluster.  Of note, the four Shakespearian clusters marked with a red circle are those commonly attributed to Shakespeare.  Further, the ellipses are our visual clustering assignment.

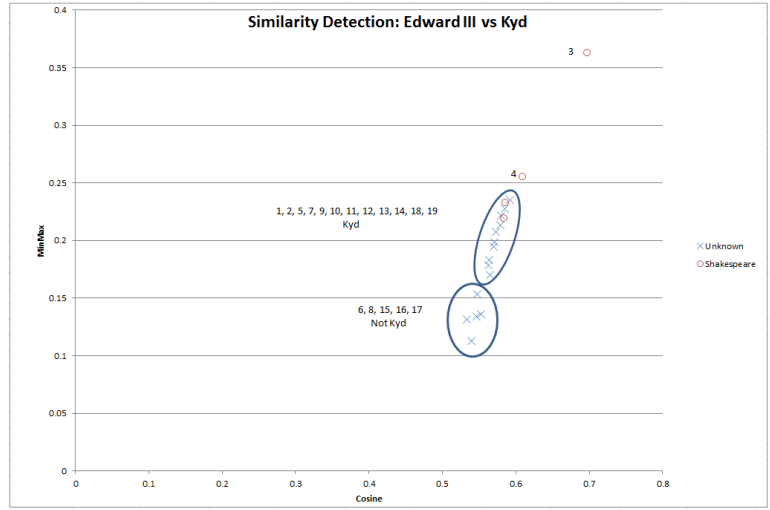

Figure 4: Using the 9-dimensional RPAS vector we compare William Shakespeare's Venus and Adonis to Edward III scene chunks using min max and cosine similarity detection. We see the 'Shakespeare 5' (chunks 6, 8, 15, 16, 17) appear in the top cluster closest to Shakespeare's work, but with the inclusion of cluster 11. The cluster in the lower left corner clearly highlights Kyd's work as different from Shakespeare. Of note, the four Shakespearian clusters marked with a red circle are those commonly attributed to Shakespeare, and none falls close to Shakespeare. Further, the ellipses are our visual clustering assignment.

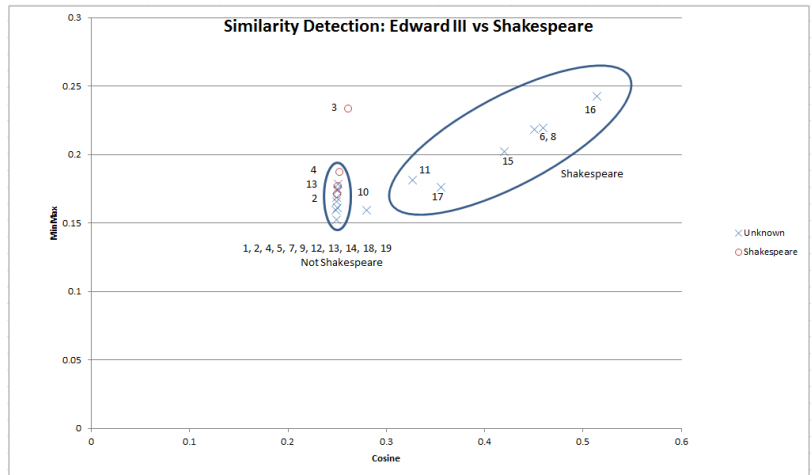

Figure 5: Using the 9-dimensional RPAS vector we use the Imposter Method and compare Christopher Marlowe's Hero and Leander to Edward III scene chunks using min max and cosine similarity detection. Marlowe's work is dissimilar to Kyd's, and therefore, the work furthest away from Marlowe's is Kyd's. Logically, if there are only two authors in Edward III, then the work closest to Marlowe must be Shakespeare. We see the 'Shakespeare 5' (chunks 6, 8, 15, 16, 17) appear in the top cluster closest to Marlowe's work, but with the inclusion of cluster 11. Of note, the four Shakespearian clusters marked with a red circle are those commonly attributed to Shakespeare, and they all fall close to Kyd. Further, the ellipses are our visual clustering assignment.

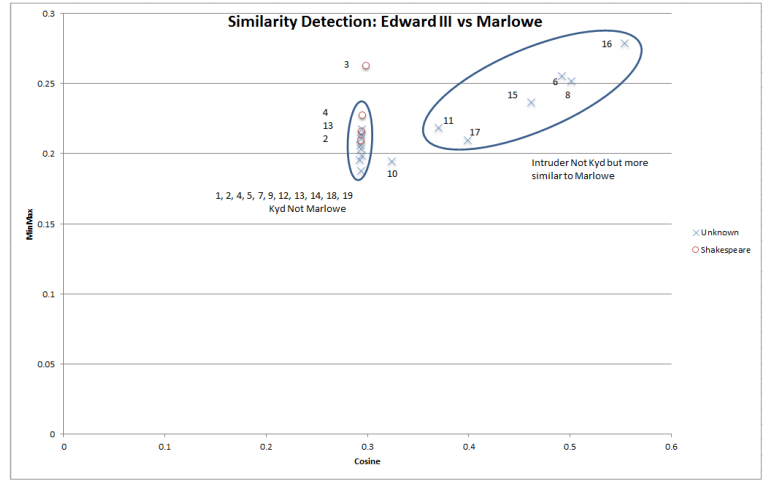

Table 1: The Seriation results of the 19 Edward III scenes show that Ordinal Leaf Ordering (OLO) technique provides the shortest Hamiltonian path. The 'Shakespeare 5' appear at the beginning distant from the commonly attributed Shakespeare scenes (marked with *).

| OLO Order | Chunk | Scene | Author |
|---|---|---|---|
| 1 | 8 | 3.4 | WS |
| 2 | 17 | 4.8 | WS |
| 3 | 15 | 4.6 | WS |
| 4 | 16 | 4.7 | WS |
| 5 | 6 | 3.2 | WS |
| 6 | 10 | 4.1 | TK |
| 7 | 9 | 4.9 | TK |
| 8 | 11 | 4.2 | TK |
| 9 | 18 | 4.5 | TK |
| 10 | 19 | 5.1 | TK |
| 11 | 13 | 4.4 | TK* |
| 12 | 14 | 4.3 | TK |
| 13 | 12 | 3.5 | TK |
| 14 | 5 | 3.1 | TK |
| 15 | 1 | 1.1 | TK |
| 16 | 2 | 1.2 | TK* |
| 17 | 7 | 3.3 | TK |
| 18 | 4 | 2.2 | TK* |
| 19 | 3 | 2.1 | TK* |

*scenes traditionally attributed to Shakespeare

Table 2: The different OLO seriation results showing changes in order when noise is added to the RPAS scene matrix. Shakespeare's work sits on the left (chunks 8, 17, 15, 16, 6), while Kyd's scenes sit on the far right with the commonly accepted Shakespeare scenes (5, 1, 2, 7, 4, 3).

| OLO Seriation of 19 Edward III scenes with introduced noise | | | | | | | | | | | | | | | | | | | Noise |
|---|---|---|---|---|---|---|---|---|---|---|---|---|---|---|---|---|---|---|---|
| 8 | 17 | 15 | 16 | 6 | 10 | 9 | 11 | 18 | 19 | 13 | 14 | 12 | 5 | 1 | 2 | 7 | 4 | 3 | 0 |
| 8 | 17 | 15 | 16 | 6 | 10 | 9 | 11 | 18 | 19 | 13 | 14 | 12 | 5 | 1 | 2 | 7 | 4 | 3 | 1 |
| 8 | 17 | 15 | 16 | 6 | 11 | 9 | 10 | 18 | 19 | 13 | 14 | 12 | 5 | 1 | 2 | 7 | 4 | 3 | 50 |
| 8 | 17 | 15 | 16 | 6 | 11 | 9 | 10 | 18 | 19 | 13 | 14 | 12 | 5 | 1 | 2 | 7 | 4 | 3 | 100 |
| 8 | 17 | 15 | 16 | 6 | 10 | 18 | 19 | 13 | 14 | 12 | 11 | 9 | 5 | 1 | 2 | 7 | 4 | 3 | 200 |
| 8 | 17 | 15 | 16 | 6 | 10 | 18 | 19 | 13 | 12 | 14 | 11 | 9 | 5 | 1 | 2 | 7 | 4 | 3 | 400 |
| 8 | 17 | 15 | 16 | 6 | 11 | 9 | 10 | 18 | 19 | 14 | 12 | 13 | 7 | 5 | 1 | 2 | 4 | 3 | 800 |
| 8 | 17 | 15 | 16 | 6 | 10 | 18 | 11 | 12 | 14 | 19 | 13 | 9 | 7 | 5 | 1 | 2 | 4 | 3 | 1000 |
| 8 | 17 | 15 | 16 | 6 | 10 | 18 | 11 | 9 | 12 | 14 | 19 | 13 | 5 | 1 | 2 | 7 | 4 | 3 | 2000 |

Table 3: The 19 scenes of the Edward III play with the Shakespearian scenes are referenced to the 19 chunks. In the third column, the five commonly attributed Shakespeare scenes are shown against those unknown authored scenes. In column four are the results of Brian Vickers' Kyd trigram scene allocation. Column five shows a summary of the results using RPAS.

| Chunk | Scene | Long-Held View | Vickers View | Our View |
|---|---|---|---|---|
| 1 | 1.1 | U | K | K |
| 2 | 1.2 | S | S | K |
| 3 | 2.1 | S | S | K |
| 4 | 2.2 | S | S | K |
| 5 | 3.1 | U | K | K |
| 6 | 3.2 | U | K | S |
| 7 | 3.3 | U | K | K |
| 8 | 3.4 | U | K | S |
| 9 | 3.5 | U | K | K |
| 10 | 4.1 | U | K | K/S* |
| 11 | 4.2 | U | K | K/S* |
| 12 | 4.3 | U | K | K |
| 13 | 4.4 | S | S | K |
| 14 | 4.5 | U | K | K |
| 15 | 4.6 | U | K | S |
| 16 | 4.7 | U | U | S |
| 17 | 4.8 | U | K | S |
| 18 | 4.9 | U | U | K |
| 19 | 5.1 | U | K | K |

* Could be a collaboration